\documentclass{article}

\usepackage{spconf}

\usepackage{graphicx}
\graphicspath{ {./figs+imgs/} }
\DeclareGraphicsExtensions{.pdf,.png,.jpg}

\usepackage{url}
\usepackage{amsmath}
\usepackage{amsthm,amssymb,amsfonts,mathrsfs,mathtools,amsbsy}
\usepackage{bm}
\usepackage{stmaryrd}
\usepackage{algorithmic}
\usepackage{algorithm}
\usepackage{array}
\usepackage[hidelinks]{hyperref}
\usepackage{siunitx}
\usepackage{cuted}
\usepackage{xspace}

\usepackage[inline]{enumitem}

\usepackage{caption}
\usepackage{subfig}
\usepackage{graphicx}
\usepackage{multirow}
\usepackage{multicol}
\usepackage{threeparttable}
\usepackage{balance}

\usepackage[%
backend = biber,%
maxcitenames = 2,
maxbibnames = 10,%
minbibnames = 2,%
style = ieee,%
citestyle = numeric-comp,%
bibencoding = utf8,%
datamodel = software,%
defernumbers = true,%
sortcites = true,%
doi = false,
isbn = false,%
url = true,%
eprint = true%
]{biblatex}

\usepackage{pgfplots}
\pgfplotsset{compat=newest}
\usepgflibrary[plotmarks]
\usepgfplotslibrary{fillbetween}
\usetikzlibrary{angles, arrows, arrows.meta, shapes, tikzmark, patterns}

\definecolor{brightlavender}{rgb}{0.75,0.58,0.89}
\definecolor{byzantine}{rgb}{0.74,0.2,0.64}
\definecolor{forestgreen}{rgb}{0.13,0.55,0.13}
\definecolor{brickred}{rgb}{0.8,0.25,0.33}
\definecolor{royalblue}{rgb}{0.25,0.41,0.88}
\definecolor{goldenrod}{rgb}{0.85,0.65,0.13}

\definecolor{myblue}{RGB}{31,119,180}
\definecolor{myorange}{RGB}{255,127,14}
\definecolor{mygreen}{RGB}{44,160,44}
\definecolor{myred}{RGB}{214,39,40}
\definecolor{mypurple}{RGB}{148,103,189}
\definecolor{mybrown}{RGB}{140,86,75}
\definecolor{mypink}{RGB}{227,119,194}
\definecolor{mygray}{RGB}{127,127,127}
\definecolor{myolive}{RGB}{188,189,34}

\pgfdeclarelayer{background}
\pgfdeclarelayer{fg}    
\pgfdeclarelayer{ffg}   
\pgfsetlayers{background,main,fg,ffg}  

\usepackage{thmtools}
\usepackage[unq]{unique}

\declaretheoremstyle[%
  headfont=\normalfont\bfseries,
  notefont=\mdseries,
  notebraces={(}{)},
  bodyfont=\normalfont,
  postheadspace=1ex
]{mystyle}

\declaretheorem[style=mystyle,
  name=Modeling~Assumptions,
  refname={assumptions},
  Refname={Assumptions},
]{assumptions}

\declaretheorem[style=mystyle,
  name=Definition,
  refname={definition,definitions},
  Refname={definition,definitions},
]{definition}

\declaretheorem[style=mystyle,
  name=Fact,
  refname={fact,facts},
  Refname={fact,facts},
]{fact}

\newlist{assslist}{enumerate}{1}
\setlist[assslist]{label=\textbf{(\roman{*})}, ref=\theassumptions(\roman{*}), noitemsep}

\usepackage[capitalize]{cleveref}

\crefname{thm}{Theorem}{Theorems}
\crefname{prop}{Proposition}{Propositions}
\crefname{assumptions}{Modeling~Assumptions}{Modeling~Assumptions}
\crefname{lemma}{Lemma}{Lemmata}
\crefname{definition}{Definition}{Definitions}
\crefname{example}{Example}{Examples}
\crefname{algo}{Algorithm}{Algorithms}
\crefname{fact}{Fact}{Facts}
\crefname{claim}{Claim}{Claims}
\crefname{appendix}{Appendix}{Appendices}
\crefname{coroll}{Corollary}{Corollaries}
\crefname{figure}{Figure}{Figures}
\crefname{section}{Section}{Sections}

\crefname{thmlisti}{Theorem}{Theorems}
\crefname{lemlisti}{Lemma}{Lemmata}
\crefname{proplisti}{Proposition}{Propositions}
\crefname{asslisti}{Modeling~Assumption}{Modeling~Assumptions}
\crefname{assslisti}{Modeling~Assumption}{Modeling~Assumptions}
\crefname{deflisti}{Definition}{Definitions}
\crefname{exlisti}{Example}{Examples}
\crefname{algolisti}{Algorithm}{Algorithms}
\crefname{factlisti}{Fact}{Facts}
\crefname{claimlisti}{Claim}{Claims}
\crefname{applisti}{Appendix}{Appendices}
\crefname{MyEnumSeci}{}{}
\crefname{MyEnumSubSeci}{}{}

\colorlet{tableMulti}{red!20}
\colorlet{tableSingle}{cyan!30}

\newcommand{\IntegerPP}{\mathbb{N}_*}
\newcommand{\Real}{\mathbb{R}}

\newcommand{\RealPP}{\mathbb{R}_{++}}

\newcommand\given{{\mathbin{}\mid\mathbin{}}}
\newcommand\vect[1]{\mathbf{#1}}
\newcommand\vectcal[1]{\mathbfcal{#1}}

\newcommand{\samp}{\mathop{P_{\Omega}}}

\DeclareMathOperator{\rank}{rank}
\DeclareMathOperator{\ranktt}{{rank}_{TT}}
\DeclareMathOperator{\cranktt}{crank_{TT}}
\DeclareMathOperator{\tovec}{vec}
\DeclareMathOperator{\grad}{grad}

\DeclarePairedDelimiter\norm{\lVert}{\rVert}
\DeclarePairedDelimiterX\innerp[2]{\langle}{\rangle}{#1
  \mathop{}\delimsize\vert\mathop{} #2}

\providecommand\given{} 
\newcommand\SetSymbol[1][]{
  \nonscript\,#1\vert \allowbreak \nonscript\,\mathopen{}}
\DeclarePairedDelimiterX\Set[1]{\lbrace}{\rbrace}%
{ \renewcommand\given{\SetSymbol[\delimsize]} #1 }

\DeclareMathAlphabet\mathbfcal{OMS}{cmsy}{b}{n}
\DeclareMathAlphabet\mathbfit{OML}{cmm}{b}{it}
\DeclareMathAlphabet\mathbfscr{OMS}{mdugm}{b}{n}

\DeclarePairedDelimiter{\ceil}{\lceil}{\rceil}

\makeatletter
\newcommand*{\ie}{%
  \@ifnextchar{,}%
  {i.e.}%
  {i.e.,\@\xspace}%
}
\newcommand*{\eg}{%
  \@ifnextchar{,}%
  {e.g.}%
  {e.g.,\@\xspace}%
}
\newcommand*{\cf}{%
  \@ifnextchar{,}%
  {cf.}%
  {cf.,\@\xspace}%
}
\makeatother

\usepackage[mathlines, switch]{lineno}

\newcommand*\patchAmsMathEnvironmentForLineno[1]{%
  \expandafter\let\csname old#1\expandafter\endcsname\csname #1\endcsname
  \expandafter\let\csname oldend#1\expandafter\endcsname\csname end#1\endcsname
  \renewenvironment{#1}%
  {\linenomath\csname old#1\endcsname}%
  {\csname oldend#1\endcsname\endlinenomath}}%
\newcommand*\patchBothAmsMathEnvironmentsForLineno[1]{%
  \patchAmsMathEnvironmentForLineno{#1}%
  \patchAmsMathEnvironmentForLineno{#1*}}%
\AtBeginDocument{%
  \patchBothAmsMathEnvironmentsForLineno{equation}%
  \patchBothAmsMathEnvironmentsForLineno{align}%
  \patchBothAmsMathEnvironmentsForLineno{flalign}%
  \patchBothAmsMathEnvironmentsForLineno{alignat}%
  \patchBothAmsMathEnvironmentsForLineno{gather}%
  \patchBothAmsMathEnvironmentsForLineno{multline}%
}

\usepackage{stackengine}
\stackMath
\newlength\matfield
\newlength\tmplength

\newcommand{\krettah}{\textnormal{KReTTaH}\xspace}

\title{Kernel Regression of Multi-Way Data via Tensor Trains with\\ Hadamard Overparametrization: The Dynamic Graph Flow
Case
\vspace{-10pt}}

\name{Duc Thien Nguyen$^{\star}$\thanks{The work of D.~T.~Nguyen was supported by JST SPRING, Japan Grant Number
    JPMJSP2180. The work of E.~Kofidis has been partly supported by the University of Piraeus Research Center. The work
    of D.~A.~Pados was supported by the US Air Force Office of Scientific Research under Grant W911NF-20-1-028.} \qquad
  Konstantinos Slavakis$^{\star}$ \qquad Eleftherios Kofidis$^{\dagger}$ \qquad Dimitris
  Pados$^{\sharp}$ \vspace{-16pt}}

\address{\small$^{\star}$Department of Information and Communications Engineering, Institute of Science Tokyo,
  Japan \vspace{-5pt}\\
  \small (emails: \texttt{duc.t.45ae@m.isct.ac.jp, slavakis@ict.eng.isct.ac.jp}) \\
  \small$^{\dagger}$Department of Statistics and Insurance Science, University of Piraeus, Greece (email:
  \texttt{kofidis@unipi.gr}) \\
  \small$^{\sharp}$CA-AI, Department of Electrical Engineering and Computer Science, Florida Atlantic University, USA
  (email: \texttt{dpados@fau.edu}) \vspace{-16pt}
}

\begin{document}

\ninept
\sloppy
\maketitle

\begin{abstract}
  A regression-based framework for interpretable multi-way data imputation, termed Kernel Regression via Tensor Trains
  with Hadamard overparametrization (\krettah), is introduced. \krettah adopts a nonparametric formulation by casting
  imputation as regression via reproducing kernel Hilbert spaces. Parameter efficiency is achieved through tensors of fixed tensor-train (TT) rank, which reside on low-dimensional Riemannian manifolds, and is further enhanced via Hadamard
  overparametrization, which promotes sparsity within the TT parameter space. Learning is accomplished by solving a
  smooth inverse problem posed on the Riemannian manifold of fixed TT-rank tensors. As a representative application, the
  estimation of dynamic graph flows is considered. In this setting, \krettah exhibits flexibility by seamlessly
  incorporating graph-based (topological) priors via its inverse problem formulation. Numerical tests on real-world
  graph datasets demonstrate that \krettah consistently outperforms state-of-the-art alternatives---including a
  nonparametric tensor- and a neural-network-based methods---for imputing missing, time-varying edge flows.
\end{abstract}


\vspace{-16pt}
\section{Introduction}
\vspace{-0.1cm}

Multi-way data, naturally represented as multidimensional arrays or tensors~\cite{Sidiropoulos:ieeeTSP:17}, often miss a
number of their entries due to, \eg, imperfections in data acquisition. Tensor completion (TC) aims to estimate these missing
values by leveraging correlations that arise from an underlying low-rank structure. Unlike matrices, the tensor rank is
defined for specific tensor decomposition (TD) models, which approximate a tensor using smaller (core) tensors and
matrices. The canonical polyadic decomposition (CPD) and Tucker decomposition (TKD) models are the most
popular~\cite{Sidiropoulos:ieeeTSP:17}. Although being the simplest and enjoying mild uniqueness conditions, CPD may be
too restrictive for some datasets. On the other hand, TKD provides enhanced expressiveness, albeit at higher
computational and memory costs.

An intermediate alternative is the tensor-train (TT) model~\cite{oseledets2011tensor} (and its variants, such as the
tensor ring decomposition (TRD)~\cite{zhao2016tensor}). Compared with CPD, TT decomposition (TTD) improves feature extraction,
accuracy, and stability, while its core ranks are easier to determine~\cite{TTthesis2022}. Relative to TKD, TTD achieves
greater compactness and mitigates the curse of dimensionality by using only low-order ($\leq 3$) core
tensors~\cite{vanloan2015}. Remarkably, all tensors of fixed TT-rank exhibit a rich geometric structure, forming
low-dimensional Riemannian manifolds embedded within high-dimensional ambient spaces~\cite{holtz2012manifolds,
RobbinSalamon:22}. This latent geometry can be exploited for both analytical and computational purposes.

Although TD models have witnessed success in TC~\cite{song2019}, their ``blind'' decomposition often underuse side information, which can be crucial for imputation. Such prior knowledge can be incorporated via kernel methods, which map data into high-dimensional feature spaces and thereby capture complex nonlinear dependencies~\cite{aronszajn1950theory, scholkopf2002learning, kimeldorf1971some}. Kernels have been incorporated into several tensor models such as the CPD~\cite{bazerque2012nonparametric}, TKD~\cite{zhao2013kernel}, TTD~\cite{gorodetsky2018gradient}, and TRD~\cite{huang2025kernel} to encode side information or serve as implicit basis functions.
However, methods based on CPD and TKD~\cite{bazerque2012nonparametric,zhao2013kernel} may face limitations, \eg, the rigid representation of CPD or the scaling problem of TKD. Meanwhile, the parametric framework~\cite{huang2025kernel} assumes Gaussian distribution for the observed data as well as the entries of the TRD's core tensors.

This paper introduces an \textit{interpretable}\/ TT-based framework for multi-way data regression, termed
\textit{Kernel Regression via TT with Hadamard overparametrization (\krettah).} 
\krettah formulates regression via
reproducing kernel Hilbert space (RKHSs)~\cite{aronszajn1950theory, scholkopf2002learning, kimeldorf1971some}, enabling
nonparametric and nonlinear functional approximation.
Extending previous work~\cite{multilkrim, nguyen2025imputation,
nguyen:apsipa25} to multi-way data, \krettah adapts TTs into its parameter tensors for a novel bi-tensorial modeling (see~\eqref{eq:model}).
While~\cite{multilkrim, nguyen2025imputation} adopt an implicit manifold learning approach with
user-unknown underlying manifolds, \krettah utilizes an explicit construction, specifying the parameter space as
user-defined TT-rank Riemannian manifolds~\cite{holtz2012manifolds}.
To enhance approximation capabilities,
Hadamard overparametrization (HP) is applied to the parameter tensors of fixed TT-rank, a feature unexplored in kernel-based TTD~\cite{gorodetsky2018gradient}. In fact, when combined with smooth regularizers in the learning
objective, HP---contrary to intuition---promotes sparsity and induces further dimensionality
reduction in the parameter space, as evidenced by~\cite{hoff2017lasso, li2023tail, ziyin2023spred,
  kolb2024smoothing}. Via the explicit manifold structure and the sparsity-inducing effect of
overparametrization, \krettah yields a smooth inverse problem, cast on Riemannian manifolds of fixed TT-rank,
which can be solved using any Riemannian optimization method. This further distinguishes \krettah from~\cite{bazerque2012nonparametric,zhao2013kernel,gorodetsky2018gradient,huang2025kernel}, which employ alternating minimization or gradient descent.

Dynamic edge flow imputation in graphs serves as an illustrative application, demonstrating the ability of \krettah to
seamlessly incorporate graph-based (topological)~\cite{giblin:10, lim2020hodge, schaub2021signal,
  barbarossa2020topological, battiston2020networks} priors into its inverse problem formulation. In contrast to
classical ``black-box'' neural-network (NN) methods, \krettah provides an \textit{interpretable}\/ solution through
regression in RKHSs, geometric reasoning in Riemannian manifolds, and sparse coding.
Numerical tests on real-world graph datasets show that \krettah outperforms leading methods for imputing missing
time-varying edge flows, including a nonparametric CPD and a NN~\cite{bazerque2012nonparametric,
steinlechner2016riemannian, tt-AdaliGroup:21, yu2025robust, rodd2019hodgenet, roddenberry2023signal,
krishnan2024simplicial, money2024evolution}.
Additional numerical tests, such as performance of~\cite{zhao2013kernel,gorodetsky2018gradient,huang2025kernel} will be reported in a future work.

\vspace{-0.2cm}
\section{Preliminaries}\label{sec:prem}
\vspace{-0.3cm}

\subsection{Basic notation}

For an order-$N$ tensor $\vectcal{X}\in \Real^{I_1\times I_2\times \cdots \times I_N}$, $\vectcal{X}({i_1,i_2,
  \ldots,i_N})$ denotes its $({i_1,i_2, \ldots,i_N})$th entry, with $i_j \in \llbracket 1, I_j\rrbracket \coloneqq
\Set{1, \ldots, I_j}$ and $j\in \llbracket 1, N\rrbracket$. The vectorization of $\vectcal{X}$---according to a
user-defined ordering of the tensor indices---yields the $(I_1 I_2 \cdots I_N) \times 1$ $\tovec ( \vectcal{X}
)$. Fixing one index while varying all others yields a ``slice'' of $\vectcal{X}$, \eg, $\vectcal{X}(:, \ldots, :, i)$,
where $\colon$ denotes a full range of indices.
The $m$th unfolding of $\vectcal{X}$ is obtained by arranging its entries in a matrix \( \vectcal{X}^{\langle m\rangle}
\in \Real^{(I_1 \cdots I_m) \times (I_{m+1} \cdots I_N)} \).
The Hadamard (entry-wise) product of two same-sized tensors
$\vectcal{X}$ and $\vectcal{Y}$ is written as $\vectcal{X}\odot \vectcal{Y}$.
The inner product between two same-sized tensors
$\vectcal{X}$ and $\vectcal{Y}$ is $\innerp{\vectcal{X}}{\vectcal{Y}} \coloneqq \tovec^{\intercal}(\vectcal{X})\,
\tovec(\vectcal{Y})$, where $^{\intercal}$ stands for vector/matrix transposition. The Frobenius norm of $\vectcal{X}$ is
defined by $\norm{\vectcal{X}}_\textnormal{F} \coloneqq \innerp{\vectcal{X}} {\vectcal{X}}^{1/2}$,
while its $\ell_p$-norm as $\norm{\vectcal{X}}_{p} \coloneqq (\sum_{i_1, i_2, \ldots, i_N} \lvert \vectcal{X}(i_1, i_2,
\ldots, i_N)\rvert^p )^{1/p}$. 
For an index set $\Omega \subset \llbracket 1, I_1 \rrbracket \times \ldots \times
\llbracket 1, I_N \rrbracket$, the ``subtensor'' $\vectcal{Y}_{\Omega}$ consists of all entries indexed by $\Omega$. 
The sampling mapping $\samp
\colon \Real^{I_1 \times \cdots \times I_N} \to \Real^{I_1 \times \cdots \times I_N}$ is defined as follows:
$\samp( \vectcal{Y} )(i_1,\ldots,i_N) \coloneqq \vectcal{Y}(i_1, \ldots, i_N)$ if $(i_1,\ldots,i_N) \in \Omega$, and
$\samp(\vectcal{Y})(i_1,\ldots,i_N) \coloneqq 0$ otherwise.
For convenience, let $\IntegerPP$ and $\RealPP$ be the
sets of positive integers and positive real numbers, respectively.

\vspace{-0.2cm}
\subsection{Example: Dynamic graph flows as multi-way data}\label{sec:prem.sc}

This subsection sets the stage for the discussion in~\cref{sec:inv.problem,sec:tests} by formulating the edge-flow
signals as multi-way data.

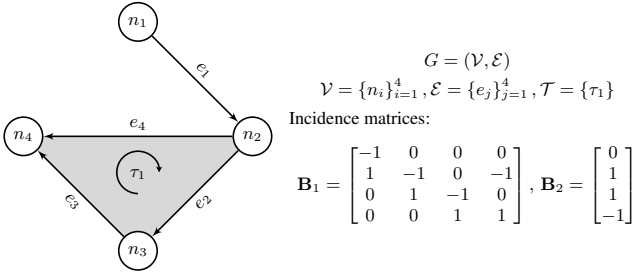
\begin{figure}
  \centering \resizebox{\linewidth}{!}{\begin{minipage}{0.65\textwidth}
\begin{tikzpicture}[>=latex',thick]

    \node (v1) at (2,2) [draw, circle, fill=white] {$n_1$};
    \node (v2) at (4,0) [draw, circle, fill=white] {$n_2$};
    \node (v3) at (2,-2) [draw, circle, fill=white] {$n_3$};
    \node (v4) at (0,0) [draw, circle, fill=white] {$n_4$};

    \draw[->] (v1) edge node [black,sloped,anchor=south] {$e_1$} (v2);
    \draw[->] (v2) edge node [black,sloped,below] {$e_2$} (v3);
    \draw[->] (v3) edge node [black,sloped,below] {$e_3$} (v4);
    \draw[->] (v2) edge node [black,sloped,anchor=south] {$e_4$} (v4);

    \node at (barycentric cs:v2=1,v3=1,v4=1) {$\tau_1$
    };

    \draw[-{Stealth[length=3pt, width=3pt]}, thick] (barycentric cs:v2=1,v3=2,v4=1) arc (-90:-360:0.37cm);

    \begin{pgfonlayer}{background}
        \fill[fill=lightgray, fill opacity=0.6] (v2.center) to (v3.center) to (v4.center);
    \end{pgfonlayer}

\end{tikzpicture}
\end{minipage}%
\hspace{-6.66cm}
\begin{minipage}{0.35\textwidth}

\[
G=(\mathcal{V}, \mathcal{E})
\]
\[
\mathcal{V}=\Set{n_i}_{i=1}^4 \,, 
\mathcal{E}=\Set{e_j}_{j=1}^4 \,,
\mathcal{T}=\{\tau_1\}
\]

Incidence matrices:
\[
\vect{B}_1 = \begin{bmatrix}
-1 & 0 & 0 & 0 \\
1 & -1 & 0 & -1 \\
0 & 1 & -1 & 0\\
0 & 0 & 1 & 1
\end{bmatrix}, \,
\vect{B}_2 = \begin{bmatrix}
0 \\
1 \\
1 \\
-1 \\
\end{bmatrix} 
\]


\end{minipage}}
  \vspace{-7pt}
  \caption{An example of a graph, $G=(\mathcal{V}, \mathcal{E})$, and its set of triangles, $\mathcal{T}$.}
  \label{fig:sc.example}
\end{figure}

A graph $G = (\mathcal{V}, \mathcal{E})$ is defined by a set of nodes $\mathcal{V}$ and a set of edges $\mathcal{E}
\subseteq \mathcal{V} \times \mathcal{V}$. Let $\mathcal{T}$ be the set of its triangles, defined as fully connected
subgraphs, each comprising three nodes. Let $N_0 = \lvert \mathcal{V} \rvert$ be the number of nodes, $N_1 =
\lvert\mathcal{E}\rvert$ the number of edges, and $N_2 = \lvert \mathcal{T} \rvert$ the number of triangles of
$G$~\cite{giblin:10, lim2020hodge,schaub2021signal, barbarossa2020topological}.  The incidence matrix $\vect{B}_1 \in
\Real^{N_0 \times N_1}$ captures the node-to-edge adjacencies, while $\vect{B}_2 \in \Real^{N_1 \times N_2}$ encodes the
edge-to-triangle ones. These incidence matrices depend on an arbitrary choice of orientation: an edge $e$ is oriented as
an ordered pair of nodes $(i,j)$, and a triangle $\tau$ as $(m,n,p)$. Accordingly, the entries of the node-to-edge
incidence matrix are $\vect{B}_1(i,e) = -\vect{B}_1(j,e) = -1$, with $\vect{B}_1(k,e) = 0$ for all $k \neq i, j$. In the
incidence matrix $\vect{B}_2$, the entry $\vect{B}_2(e,\tau)$ is zero unless $e$ is an edge of $\tau$; it equals~1 if
$e$ and $\tau$ share the same orientation and -1 otherwise. An example of a graph $G$ with its incidence matrices are
shown in~\cref{fig:sc.example}.  In general, time-varying flows are associated with graph edges.

Time-varying edge-flow signals, such as traffic or communication flows, are arranged in a tensor $\vectcal{Y} \in
\Real^{I_1 \times I_2 \times I_3}$, where $\vectcal{Y}(i_1,i_2,i_3)$ denotes the flow on edge $i_1$ at the $i_2$th time
point of the $i_3$th time interval. Here, $I_1 = N_1$ is the number of edges, $I_2$ the number of time points per
interval (\eg, hours per day), and $I_3$ the number of intervals (\eg, days). Each column of the 1st unfolding
$\vectcal{Y}^{\langle 1 \rangle} \in \Real^{I_1 \times I_2 I_3}$ thus represents a snapshot of flows at a particular
time $t \in \llbracket 1, I_2 I_3 \rrbracket$. Edge flows are typically assumed to be almost divergence-free, \ie,
nearly conserved at nodes, which can be expressed as $\norm{ \vect{B}_1 \vectcal{Y}^{\langle 1 \rangle} }_{\mathrm{F}}
\approx 0$, and curl-free, \ie, summing to nearly zero around triangles, that is, $\norm{ \vect{B}_2^{\intercal}
  \vectcal{Y}^{\langle 1 \rangle} }_{\mathrm{F}} \approx 0$~\cite{battiston2020networks, barbarossa2020topological,
  schaub2021signal}.

\vspace{-0.2cm}
\subsection{Tensor-train tensors and their Riemannian geometry}

\begin{definition}[Mode-$(N,1)$ contraction~\cite{cichocki2016tensor}]\label{def:mode1contraction}
  For $\vectcal{X}_1 \in \Real^{I_1\times \cdots \times I_N}$ and $\vectcal{X}_2 \in \Real^{J_1\times \cdots \times
    J_M}$, with $I_N = J_1$, the mode-$(N,1)$ contraction $\times_N^1$ yields $\vectcal{Z} \coloneqq \vectcal{X}_1
  \times_N^1 \vectcal{X}_2 \eqqcolon \vectcal{X}_1 \times^1 \vectcal{X}_2 \in \Real^{I_1\times \cdots \times
    I_{N-1}\times J_2\times \cdots \times J_M}$, with $\vectcal{Z} (i_1, \ldots, i_{N-1}, j_2, \ldots, j_M) \coloneqq
  \sum_{i_N \in \llbracket 1, I_N \rrbracket} \vectcal{X}_1(i_1, \ldots, i_N)\, \vectcal{X}_2(i_N, j_2, \ldots, j_M)$.
\end{definition}

\begin{definition}[TT-tensors and TTD~\cite{oseledets2011tensor}]\label{def:tt.format}
  Given order-3 \textit{core} tensors $\vectcal{A}_k \in \Real^{r_{k-1}\times I_k \times r_k}$, $\forall k \in
  \llbracket 1, N\rrbracket$, with $r_0=r_N=1$ and $\vect{r} \coloneqq (r_0, r_1, \ldots, r_{N-1}, r_N) \in
  \IntegerPP^{N+1}$, the associated TT-tensor is defined as the $I_1 \times \cdots \times I_N$ tensor $\vectcal{A}
  \coloneqq \vectcal{A}_1 \times^1 \vectcal{A}_2 \times^1 \cdots \times^1 \vectcal{A}_N$. It can be verified that
  $\vectcal{A} (i_1,\ldots,i_N) = \vectcal{A}_1 (1, i_1, :) \vectcal{A}_2 (:, i_2, :) \cdots \vectcal{A}_N (:, i_N,
  1)$~\cite{oseledets2011tensor}; notice that $\vectcal{A}_1 (1, i_1, :)$ and $\vectcal{A}_N (:, i_N, 1)$ are row and
  column vectors, respectively. The vector $\vect{r}$ is often called the TT compression rank of $\vectcal{A}$, and is
  denoted hereafter by $\cranktt ( \vectcal{A} )$. For any tensor $\vectcal{X}\in \Real^{I_1 \times \cdots \times I_N}$,
  and given an $\vect{r} \coloneqq (1, r_1, \ldots, r_{N-1}, 1) \in \IntegerPP^{N+1}$, the TT-decomposition (TTD) of
  $\vectcal{X}$ is defined as the TT-tensor $\vectcal{A}_*$ which best approximates $\vectcal{X}$ in the Frobenius norm
  sense and under the constraint $\cranktt (\vectcal{A}_*) \preceq \vect{r}$, where $\preceq$ stands for entry-wise
  $\leq$.
\end{definition}

\begin{definition}[TT-rank~\cite{oseledets2011tensor}]
  The TT-rank of a tensor $\vectcal{X}$ is defined as the function $\ranktt (\cdot) \colon \Real^{I_1 \times \cdots \times I_N}
  \to \IntegerPP^{N+1} \colon \vectcal{X} \mapsto \ranktt (\vectcal{X}) \coloneqq (1, \rank(\vectcal{X}^{\langle
    1\rangle}), \ldots, \rank (\vectcal{X}^{\langle N-1\rangle}), 1)$.
\end{definition}


\begin{fact}[{\cite[Thm.~2.1]{oseledets2011tensor}}]
  Any $\vectcal{X} \in \Real^{I_1 \times \cdots \times I_N}$ has a TTD $\vectcal{A}_*$ with $\cranktt( \vectcal{A}_* ) \preceq
  \ranktt(\vectcal{X})$, which can be computed by the tensor-train singular value decomposition
  (TT-SVD)~\cite[Alg.~1]{oseledets2011tensor}.
\end{fact}

For an $\vect{r} \coloneqq (1, r_1, \ldots, r_{N-1}, 1) \in \IntegerPP^{N+1}$, the set of all tensors with TT-rank equal
to $\vect{r}$, that is, $\mathcal{M}_{\vect{r}} \coloneqq \{ \vectcal{X}\in \Real^{I_1\times \cdots \times I_N} \given
\ranktt(\vectcal{X})=\vect{r} \}$, is a Riemannian manifold of dimension $\sum_{k=1}^N r_{k-1}I_k r_k-\sum_{k=1}^{N-1}
r_k^2$~\cite{holtz2012manifolds}. Necessary and sufficient conditions for $\mathcal{M}_{\vect{r}}$ to be nonempty are
$r_{k-1}\leq I_k r_k$ and $r_k \leq I_kr_{k-1}$, $\forall k \in \llbracket 1,N
\rrbracket$~\cite[(9.32)]{uschmajew2020geometric}.

By denoting the tangent space of $\mathcal{M}_{\vect{r}}$ at $\vectcal{X}$ as $T_\vectcal{X} \mathcal{M}_{\vect{r}}$,
the Riemannian metric is given as $\innerp{\bm{\xi}}{\bm{\eta}}_\vectcal{X} \coloneqq \innerp{\bm{\xi}}{\bm{\eta}}$,
where the tangent vectors $\bm{\xi}, \bm{\eta} \in T_{\vectcal{X}} \mathcal{M}_{\vect{r}}$ are seen as tensors in the
ambient space, $\Real^{I_1\times \cdots \times I_N}$~\cite{RobbinSalamon:22}. Since $\mathcal{M}_{\vect{r}}$ is embedded
in $\Real^{I_1 \times \cdots \times I_N}$, the Riemannian gradient of a smooth function $\mathcal{L} \colon
\mathcal{M}_{\vect{r}} \to \Real$ is given as the orthogonal projection of the classical gradient onto the tangent
space, \ie, $\grad \mathcal{L} (\vectcal{X}) \coloneqq P_{T_\vectcal{X} \mathcal{M}_{\vect{r}}}(\nabla_\vectcal{X}
\mathcal{L} (\vectcal{X}))$~\cite{absil2008manifold}. A retraction $R_{\vectcal{X}} (\cdot) \colon T_\vectcal{X}
\mathcal{M}_{\vect{r}} \to \mathcal{M}_{\vect{r}}$ maps a tangent vector back to the
manifold~\cite{absil2008manifold}. In the TT context, $R_\vectcal{X} (\cdot)$ is computable by TT-SVD~\cite{holtz2012manifolds}.

\vspace{-0.2cm}

\section{Proposed method}\label{sec:modeling}

\vspace{-0.1cm}

\subsection{Data modeling}

The $I_1 \times \ldots \times I_N$ tensor $\vectcal{Y}$ denotes the ground-truth $N$-way data, with $\Omega$ indicating
the observed entries. For the user-defined index sets $\Set{ \Omega_n }_{n=1}^{ N_{\text{nav}} }$, consider subtensors
$\Set{ \vectcal{Y}_{ \Omega_n } }_{n=1}^{ N_{\text{nav}} }$ of the observed data $\samp(\vectcal{Y})$, often called
navigator data.  For instance, in \num{2}-way image data $\vectcal{Y}$, these index sets may correspond to image
patches. Structural information within these navigator data is extracted next and incorporated into the proposed data
model. Without loss of generality, assume all $\Set{ \Omega_n }$ have the same cardinality $\nu$, typically much smaller
than the total number of entries in $\vectcal{Y}$. For convenience, let $\vect{y}_n \coloneqq \tovec ( \vectcal{Y}_{
  \Omega_n } ) \in \Real^{\nu}$. To reduce the computational burden when $N_{\text{nav}}$ is very large, a subset $\Set{
  \mathbfit{l}_k }_{k=1}^{N_{\mathit{l}}}$, referred to as landmark points~\cite{de2004sparse}, with $N_{\mathit{l}}
\leq N_{\text{nav}}$, is selected from $\Set{\vect{y}_n}_{n=1}^{N_\text{nav}}$ by a user-defined strategy.

To effect kernel-based approximation, a feature map $\varphi(\cdot)$ transforms each landmark point to
$\varphi(\mathbfit{l}_k) \coloneqq \kappa(\mathbfit{l}_k, \cdot)$ within an RKHS $\mathscr{H}$ with reproducing kernel
$\kappa(\cdot, \cdot) \colon \Real^{\nu}\times \Real^{\nu} \to \Real$ (for example, $\kappa$ may be the celebrated
Gaussian function~\cite{aronszajn1950theory, kimeldorf1971some, scholkopf2002learning}). The kernel matrix
$\vect{K}_{\vectcal{Y}_{\Omega} }$ is then defined as the $N_{\mathit{l}} \times N_{\mathit{l}}$ matrix with entries
$\vect{K}_{\vectcal{Y}_{\Omega} } (k, k^{\prime}) \coloneqq
\innerp{\varphi(\mathbfit{l}_k)}{\varphi(\mathbfit{l}_{k^{\prime}})}_{\mathscr{H}} =\kappa(\mathbfit{l}_k,
\mathbfit{l}_{k^{\prime}})$, where $\innerp{\cdot}{\cdot}_{\mathscr{H}}$ denotes the inner product in $\mathscr{H}$, and
the last equality follows from the reproducing property of $\kappa$~\cite{aronszajn1950theory, kimeldorf1971some,
  scholkopf2002learning}. Loosely speaking, $\vect{K}_{\vectcal{Y}_{\Omega} }$ encodes nonlinear correlations among the
landmark points $\Set{ \mathbfit{l}_k }_{k=1}^{N_{\mathit{l}}}$. The subscript in $\vect{K}_{\vectcal{Y}_{\Omega} }$ is
used to underline the fact that the kernel matrix carries information extracted from the observed entries
$\vectcal{Y}_{\Omega}$.

For a user-defined unfolding mode $m \in \llbracket 1,N-1 \rrbracket$, the $(i_1, i_2, \ldots, i_N)$th entry of
$\vectcal{Y}$ is approximated as
\begin{alignat}{2}
  \vectcal{Y} (i_1, i_2, \ldots, i_N)
  & {} \approx {}
  && {} f_{i_1,\ldots,i_m}( \check{\bm{\mu}}_{i_{m+1}, \ldots, i_N} ) \notag \\
     & {} = {}
     && \innerp{f_{i_1,\ldots,i_m}}{ \varphi( \check{\bm{\mu}}_{i_{m+1},\ldots,i_N}) }_{\mathscr{H}}
  \,, \label{xij.reproducing.property}
\end{alignat}
where $f_{i_1, \ldots, i_m}(\cdot) \colon \Real^{\nu} \to \Real$, taken from the RKHS $\mathscr{H}$, is a function to be
identified, $\check{\bm{\mu}}_{i_{m+1}, \ldots, i_N}$ is a $\nu \times 1$ vector to be also inferred, and the last
equality in~\eqref{xij.reproducing.property} is because of the celebrated reproducing property of
$\mathscr{H}$~\cite{aronszajn1950theory, scholkopf2002learning}. Motivated by the representer
theorem~\cite{scholkopf2002learning, kimeldorf1971some}, $f_{i_1, \ldots, i_m}$ is assumed to belong to the linear span
of $\Set{ \varphi( \mathbfit{l}_k) }_{k=1}^{ N_{\mathit{l}} }$, \ie, $f_{i_1,\ldots,i_m} = \sum_{k=1}^{N_{\mathit{l}}}
u_{i_1, \ldots, i_m, k} \varphi( \mathbfit{l}_k)$, for some learnable parameters $\Set{ u_{i_1, \ldots, i_m, k} }
\subset \Real$. Similarly, $\varphi(\check{\bm{\mu}}_{i_{m+1},\ldots,i_N}) = \sum_{k^{\prime} = 1}^{N_{\mathit{l}}}
v_{k^{\prime}, i_{m+1}, \ldots, i_N} \varphi( \mathbfit{l}_{k^{\prime}} )$, for some also learnable parameters $\Set{
  v_{k^{\prime}, i_{m+1}, \ldots, i_N} } \subset \Real$, so that the linear property of the inner product in
\eqref{xij.reproducing.property} yields the kernel-based regression $\vectcal{Y} (i_1, i_2, \ldots, i_N) \approx
\sum_{k, k^{\prime}} u_{i_1, \ldots, i_m, k}\, v_{k^{\prime}, i_{m+1}, \ldots, i_N} \kappa( \mathbfit{l}_{k},
\mathbfit{l}_{k^{\prime}})$. Towards concise data modeling, define tensors $\vectcal{U} \in \Real^{I_1\times \cdots
  \times I_m \times N_{\mathit{l}}}$ and $\vectcal{V} \in \Real^{N_{\mathit{l}} \times I_{m+1}\times \cdots \times
  I_N}$, with $\vectcal{U}(i_1, \ldots, i_m, k) \coloneqq u_{i_1, \ldots, i_m, k}$ and $\vectcal{V}(k, i_{m+1}, \ldots,
i_N) \coloneqq v_{k, i_{m+1}, \ldots, i_N}$, $\forall k\in \llbracket 1, N_{\mathit{l}} \rrbracket$.

\begin{assumptions}\label{modeling.asss}\mbox{}
  \begin{assslist}

  \item\label{assume:HO} Tensors $\vectcal{U}$ and $\vectcal{V}$ are sparse. To this end, they are overparametrized via
    the Hadamard product as $\vectcal{U} \coloneqq \odot_{p=1}^{P} \vectcal{U}_p$ and $\vectcal{V} \coloneqq
    \odot_{q=1}^{Q} \vectcal{V}_q$ by the learnable tensors $\Set{ \vectcal{U}_p }_{ p = 1}^P \subset \Real^{ I_1\times
      \cdots \times I_m \times N_{\mathit{l}} }$ and $\Set{ \vectcal{V}_q }_{ q = 1}^Q \subset \Real^{N_{\mathit{l}}
      \times I_{m+1} \times \cdots \times I_N}$.

  \item\label{assume:manifolds} For user-defined TT-ranks $\vect{r}_1, \vect{r}_2$, tensor $\vectcal{U}_p$ belongs to
    the Riemannian manifold $\mathcal{M}_{\vect{r}_1} \subset \Real^{ I_1\times \cdots \times I_m \times N_{\mathit{l}}
    }$ of TT-rank $\vect{r}_1$, $\forall p\in \llbracket 1, P \rrbracket$, and $\vectcal{V}_q$ belongs to the Riemannian
    manifold $\mathcal{M}_{\vect{r}_2} \subset \Real^{N_{\mathit{l}} \times I_{m+1} \times \cdots \times I_N}$ of
    TT-rank $\vect{r}_2$, $\forall q\in \llbracket 1, Q \rrbracket$.

  \item\label{assume:final.model} The observed data $\vectcal{Y}_{\Omega}$ serve as the regressors, through
    $\vect{K}_{\vectcal{Y}_{\Omega} }$, in the following non-linear regression model:
    \begin{align}
      \vectcal{Y} \approx \underbrace{ \left( \odot_{p=1}^{P} \vectcal{U}_p \right) }_{ \vectcal{U}
      } \times^1\, \vect{K}_{\vectcal{Y}_{\Omega} }\, \times^1 \underbrace{ \left( \odot_{q=1}^{Q} \vectcal{V}_q \right)
      }_{ \vectcal{V} } \,. \label{eq:model}
    \end{align}

  \end{assslist}
\end{assumptions}

\cref{assume:HO} is motivated by recent research~\cite{hoff2017lasso, li2023tail, ziyin2023spred, kolb2024smoothing},
which, contrary to intuition, demonstrates that the overparametrization $\vectcal{U} = \odot_{p=1}^P \vectcal{U}_p$,
when combined with \textit{smooth}\/ regularization of the associated inverse problem via the Frobenius norms of the
Hadamard factors, $\sum_{p=1}^P \norm{\vectcal{U}_p}_{\text{F}}^2$ (as in~\eqref{eq:graph.task.general}), promotes
sparsity in $\vectcal{U}$ (likewise in $\vectcal{V}$). Indeed, this approach effectively induces the non-convex and
non-smooth quasi-norm regularizer $\norm{\vectcal{U}}_{2/P}^{2/P}$ for $P > 2$, which yields sparser solutions than
classical $\ell_1$-norm methods~\cite{kolb2024smoothing}. This (counter-intuitive) strategy, namely introducing
additional parameters to promote sparsity, can simultaneously enhance accuracy in approximations by increasing the
model's degrees of freedom and its representation capacity~\cite{kolb2024smoothing}.

\cref{assume:manifolds} enhances dimensionality reduction by constraining the factor tensors
$\Set{\vectcal{U}_p}_{p=1}^P$ and $\Set{\vectcal{V}_q}_{q=1}^Q$ to reside in the low-dimensional manifolds
$\mathcal{M}_{\vect{r}_1}$ and $\mathcal{M}_{\vect{r}_2}$, respectively. These manifolds are chosen to utilize the rich
geometric structure and algorithmic benefits of Riemannian geometry~\cite{RobbinSalamon:22, absil2008manifold}. Relying
on manifolds of fixed TT-rank via~\eqref{eq:model} appears to be novel in the literature on kernel-based tensor
regression and functional approximation.

Lastly, \cref{assume:final.model} significantly extends the earlier work~\cite{multilkrim, nguyen2025imputation,
  nguyen:apsipa25} from matrix to tensor data. While~\cite{multilkrim, nguyen2025imputation} adopt an implicit
manifold-learning perspective, in which the underlying manifold remains unspecified, the present framework
\emph{explicitly} constrains the solution to Riemannian manifolds of fixed TT-rank tensors. Moreover,
extending~\eqref{eq:model} from the kernel matrix $\vect{K}_{ \vectcal{Y}_{\Omega} }$ to a kernel tensor
$\vectcal{K}_{\vectcal{Y}_{\Omega} }$, capable of capturing multi-way (instead of 2-way) correlations in the data,
represents a natural and promising direction for improved performance, which will be pursued in future work.  The
previous discussion was tailored to the task of TC, with the kernel matrix constructed from the observed
data $\vectcal{Y}_{\Omega}$. Nonetheless, the proposed modeling framework is sufficiently general to accommodate kernel
matrix constructions based on prior data or user-defined bases, enabling application to a wider range of learning tasks.



\vspace{-0.2cm}
\subsection{Inverse problem}\label{sec:inv.problem}

The previous discussion motivates the following inverse problem:
\begin{subequations}\label{eq:graph.task.general}
  \begin{alignat}{3}
    \min_{ \vectcal{O} }\
    && \mathcal{L} (\vectcal{O}) & {} \coloneqq {} && \tfrac{1}{2} \norm{ \samp(\vectcal{Y}) - \samp (\vectcal{X})
    }^2_{\textnormal{F}} + \mathcal{R}(\vectcal{O}) \notag \\
    &&&&& + \tfrac{\lambda_1}{2} \sum\nolimits_{p=1}^P \norm{\vectcal{U}_p}_{\textnormal{F}}^2 +
    \tfrac{\lambda_2}{2} \sum\nolimits_{q=1}^Q
    \norm{\vectcal{V}_q}_{\textnormal{F}}^2\,, \label{eq:graph.task.loss} \\
    \text{s.to} && \vectcal{X} & \coloneqq && (\odot_{p=1}^P \vectcal{U}_p) \times^1\, \vect{K}_{\vectcal{Y}_{\Omega}
    }\, \times^1 (\odot_{q=1}^Q \vectcal{V}_q) \,, \label{graph.factorize} \\ && \vectcal{O} & \coloneqq &&
    (\vectcal{U}_1, \ldots, \vectcal{U}_P, \vectcal{V}_1, \ldots, \vectcal{V}_Q) \in \mathcal{M}_{\vect{r}_1}^P \times
    \mathcal{M}_{\vect{r}_2}^Q\,, \label{cartesian.manifold}
  \end{alignat}
\end{subequations}
where $\mathcal{M}_{\vect{r}_1}^P \times \mathcal{M}_{\vect{r}_2}^Q$ constitutes a Riemannian manifold as the Cartesian
product of Riemannian manifolds~\cite{RobbinSalamon:22}. The smooth regularizer $\mathcal{R}(\vectcal{O})$ imposes prior
knowledge specific to the application domain at hand. In dynamic-graph-flow imputation, $\mathcal{R}(\vectcal{O})
\coloneqq (\lambda_l/2) \norm{\vect{B}_1 \vectcal{X}^{\langle 1 \rangle}}_{\textnormal{F}}^2 + (\lambda_u/2)
\norm{\vect{B}_2^\intercal \vectcal{X}^{\langle 1 \rangle}}_{\textnormal{F}}^2$, motivated by the discussion in
\cref{sec:prem.sc}.  The hyperparameters $\lambda_l, \lambda_u \in \RealPP$ control the flow divergence and curl,
respectively, while $\lambda_1, \lambda_2 \in \RealPP$ control the sparsity of $\vectcal{U}$ and $\vectcal{V}$.

Owing to~\cref{modeling.asss}, the objective in~\eqref{eq:graph.task.general} is smooth, thereby enabling the use of the
powerful toolbox of Riemannian optimization~\cite{absil2008manifold}. \cref{alg:multil.general}
addresses~\eqref{eq:graph.task.general} with theoretical guarantees via a Riemannian gradient-descent method equipped
with line search for accelerated convergence~\cite{absil2008manifold}. The Riemannian gradient $\grad \mathcal{L}
(\hat{\vectcal{O}}^{(n)})$ in Line~\ref{alg.step:grad} consists of the partial Riemannian gradients
$(\grad_{\vectcal{U}_1} \mathcal{L}, \ldots, \grad_{\vectcal{U}_P} \mathcal{L}, \grad_{\vectcal{V}_1} \mathcal{L},
\ldots, \grad_{\vectcal{V}_Q} \mathcal{L})(\hat{\vectcal{O}}^{(n)})$, computable as described
in~\cite{steinlechner2016riemannian}. The retraction in Line~\ref{alg.step:retraction} is computed by
TT-SVD~\cite{steinlechner2016riemannian}. The hyperparameters for Lines~\ref{alg.step:linesearch}
and~\ref{alg.step:retraction} are $\alpha \in \RealPP$, $\beta \in (0,1)$, and $\gamma \in (0,1)$.

It is worth noting that~\cref{alg:multil.general} departs from the block alternating minimization strategy commonly used
in tensor methods, which is often sensitive to initialization and may overfit due to separate block updates~\cite{chen2020tensor,
gao2024riemannian}.
On the contrary, the proposed Riemannian gradient descent (R-GD) scheme jointly updates all factors along a smooth trajectory on the product manifold, reducing the
risk of overfitting to mode-specific noise.
Line search (Line~\ref{alg.step:linesearch}) adaptively controls step sizes, ensuring stable descent and mitigating sensitivity to poor initializations.
Moreover, the proposed R-GD approach paves the way for
generalizations to stochastic R-GD, enabling efficient online learning from streaming data.




\begin{algorithm}[!ht]
  \caption{Solving~\eqref{eq:graph.task.general} via Riemannian gradient descent}\label{alg:multil.general}
  \begin{algorithmic}[1]
    \setlength{\abovedisplayskip}{2pt}%
    \setlength{\belowdisplayskip}{2pt}%

    \ENSURE Limit point $\hat{\vectcal{O}}^{(*)}$ of the sequence $(\hat{\vectcal{O}}^{(n)})_{n\in\IntegerPP}$.

    \STATE Initialize $\hat{\mathbfcal{O}}^{(0)} \in \mathcal{M}_{\vect{r}_1}^P \times \mathcal{M}_{\vect{r}_2}^Q$, $n\leftarrow 0$

    \REPEAT \label{alg.step:resume.k}


    \STATE Get the Riemannian gradient $\bm{\xi}_n \coloneqq \grad \mathcal{L} (\hat{\vectcal{O}}^{(n)})$. \label{alg.step:grad}

    \STATE Set the descent direction as $\bm{\eta}_n = -\bm{\xi}_n$.

    \STATE\label{alg.step:linesearch} Find the smallest integer $t$ such that\
    \begin{align*}
        \mathcal{L} (\hat{\vectcal{O}}^{(n)}) - \mathcal{L} (R_{\hat{\vectcal{O}}^{(n)}}(\alpha
        \beta^t \bm{\eta}_n)) \geq -\gamma \innerp{\bm{\xi}_n}{\alpha \beta^t \bm{\eta}_n} \,.
    \end{align*}

    \STATE Update by retraction $\hat{\vectcal{O}}^{(n+1)} = R_{\hat{\vectcal{O}}^{(n)}}(\alpha
    \beta^t \bm{\eta}_n)$. \label{alg.step:retraction}


    \STATE {$n \leftarrow n+1$}

    \UNTIL{$\norm{ \hat{\vectcal{X}}^{(n)} - \hat{\vectcal{X}}^{(n-1)} }_{\text{F}} / 
        \norm{\hat{\vectcal{X}}^{(n)}}_{\text{F}} < \epsilon$}\label{alg.step:stop}
  \end{algorithmic}
\end{algorithm}

\vspace{-0.2cm}
\section{Validation via Graph-Flow Imputation}\label{sec:tests}
\vspace{-0.2cm}

\begin{figure}[!ht]
  \centering \subfloat[Eastern-Massachusetts network \label{fig:plot.ema}] {\includegraphics[width =
      .5\columnwidth]{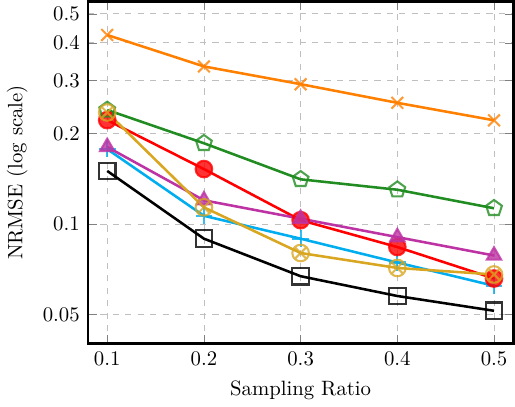}} \subfloat[Berlin-Friedrickshain network \label{fig:plot.bf}]
             {\includegraphics[width = .5\columnwidth]{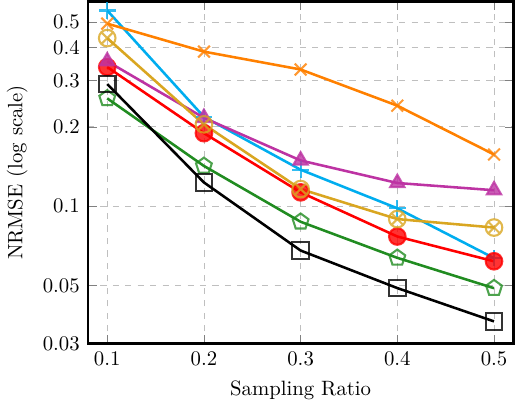}}
  \vspace{-0.1cm}
  \caption[]{Mean NRMSE value curves ($\downarrow$) vs.\ sampling
    ratios. HodgeNet~\cite{rodd2019hodgenet}:~\tikz{ \node[mark size=3pt, cyan, line width =
        1pt,]{\pgfuseplotmark{+}}; }, S-VAR~\cite{krishnan2024simplicial,
      money2024evolution}:~\tikz{ \node[mark size=3pt, orange, line width =
        1pt,]{\pgfuseplotmark{x}}; }, PS~\cite{roddenberry2023signal}:~\tikz{ \node[mark size=3pt,
        forestgreen, line width = 1pt,]{\pgfuseplotmark{pentagon}}; },
    RTTC~\cite{steinlechner2016riemannian}:~\tikz{ \node[mark size=3pt, red, line width =
        1pt,]{\pgfuseplotmark{*}}; }, STTC~\cite{yu2025robust}:~\tikz{ \node[mark size=3pt,
        byzantine, line width = 1pt,]{\pgfuseplotmark{triangle*}}; },
    NCP~\cite{bazerque2012nonparametric}:~\tikz{ \node[mark size=3pt, goldenrod, line width =
        1pt,]{\pgfuseplotmark{otimes}}; }, \krettah (proposed):~\tikz{ \node[mark size=3pt, black,
        line width = 1pt,]{\pgfuseplotmark{square}}; }~.  }
  \label{fig:nrmse}
\end{figure}

\begin{figure}[!ht]
  \centering \subfloat[Eastern-Massachusetts network \label{fig:plot.ema.sen}]
             {\includegraphics[width = .5\columnwidth]{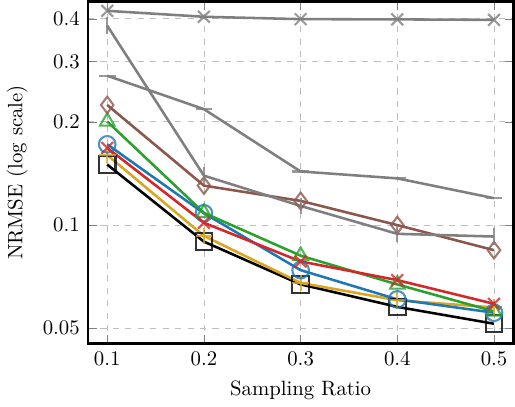}} \subfloat[Berlin-Friedrickshain
               network \label{fig:plot.bf.sen}] {\includegraphics[width =
                 .5\columnwidth]{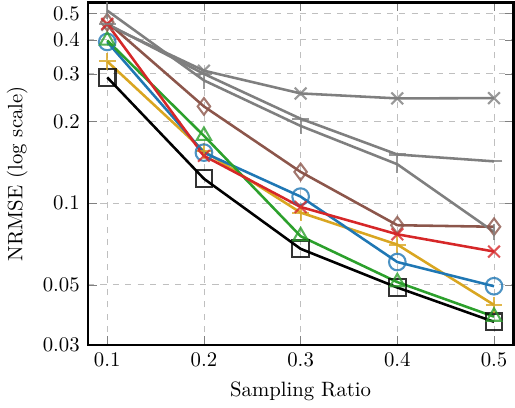}}
  \vspace{-0.1cm}
  \caption[]{The impact of different $(P,Q)$ values in \eqref{eq:model} on
    performance. $(1,1)$:~\tikz{ \node[mark size=3pt, goldenrod, line width =
        1pt,]{\pgfuseplotmark{+}}; }, $(1,3)$:~\tikz{ \node[mark size=3pt, myblue, line width =
        1pt,]{\pgfuseplotmark{o}}; }, $(2,1)$:~\tikz{ \node[mark size=3pt, mygreen, line width =
        1pt,]{\pgfuseplotmark{triangle}}; }, $(2,2)$:~\tikz{ \node[mark size=3pt, mygray, line width
        = 1pt,]{\pgfuseplotmark{x}}; }, $(2,3)$:~\tikz{ \node[mark size=3pt, mybrown, line width =
        1pt,]{\pgfuseplotmark{diamond}}; }, $(3,1)$:~\tikz{ \node[mark size=3pt, mygray, line width
        = 1pt,]{\pgfuseplotmark{|}}; }, $(3,2)$:~\tikz{ \node[mark size=3pt, myred, line width =
        1pt,]{\pgfuseplotmark{x}}; }, $(3,3)$:~\tikz{ \node[mark size=3pt, mygray, line width =
        1pt,]{\pgfuseplotmark{-}}; }, $(1,2)$:~\tikz{ \node[mark size=3pt, black, line width =
        1pt,]{\pgfuseplotmark{square}}; }~.
  }
  \label{fig:nrmse.sen}
\end{figure}

\krettah is tested on traffic-flow data for the Eastern-Massachusetts (EMA) network (\num{74} nodes, \num{258} edges,
and \num{33} triangles; see \cref{sec:prem.sc}) and the Berlin-Friedrichshain (BF) network (\num{224} nodes, \num{523}
edges, and \num{67} triangles)~\cite{transportation_networks}.  Time-varying edge-flow signals of $I_1$ edges over $I_2$
time points are generated by the traffic-flow simulator UXsim~\cite{seo2025uxsim}, which runs $I_3$ times with different
initial states and traffic volumes to generate an $I_1 \times I_2 \times I_3$ tensor $\vectcal{Y}$; $258\times 400\times
7$ for the EMA network and $523\times 350 \times 8$ for the BF network.

\krettah is compared against the state-of-the-art edge-flow imputation methods PS~\cite{roddenberry2023signal},
S-VAR~\cite{krishnan2024simplicial, money2024evolution}, and the NN-based HodgeNet~\cite{rodd2019hodgenet}, which
operate on the first unfolding of the data tensor, \ie, $\vectcal{Y}^{\langle 1 \rangle}\in \Real^{I_1 \times
I_2I_3}$. Competing tensor methods are RTTC~\cite{steinlechner2016riemannian}, a fixed-TT-rank manifold optimization method
(see also~\cite{tt-AdaliGroup:21}), STTC~\cite{yu2025robust}, a TRD for traffic-data imputation, and
NCP~\cite{bazerque2012nonparametric}, a kernel-based (nonparametric) CPD method.

The evaluation metric is the normalized root mean squared error (NRMSE), defined as $\text{NRMSE} \coloneqq
{\norm{\hat{\vectcal{X}}^{(*)} - \vectcal{Y}}_\textnormal{F}} / \norm{\vectcal{Y}}_\textnormal{F}$ (the lower the
better), where $\hat{\vectcal{X}}^{(*)} \coloneqq (\odot_{p=1}^P \hat{\vectcal{U}}_p^{(*)}) \times^1\,
\vect{K}_{\vectcal{Y}_{\Omega} }\, \times^1 (\odot_{q=1}^Q \hat{\vectcal{V}}_q^{(*)})$.  All methods are finely tuned to
reach their lowest NRMSE. Reported values are mean values of \num{10} runs with different sampled index sets and
initializations. \krettah was implemented in Python and ran on an \num{8}-core Intel(R) i7-11700 \num{2.50}~GHz CPU with
\num{32}~GB RAM.

With sampling ratio $s \in \Set{0.1, 0.2, 0.3, 0.4, 0.5}$, signals of $\ceil{I_1 \cdot s}$ edges are randomly sampled
per time instant $t \in \llbracket 1,I_2 I_3 \rrbracket$, where $\ceil{\cdot}$ is the ceiling function.  This sampling pattern
suggests that the number of observations is consistent over time. Navigator data $\Set{\vect{y}_n}_{n=1}^{N_\text{nav}}$
are formed by the columns of $\vectcal{Y}_{\Omega}^{\langle m\rangle}$ with a user-defined mode $m\in \llbracket 1,
2\rrbracket$.  Landmark points are selected via the greedy max-min-(Euclidean)-distance
strategy~\cite{de2004sparse}. Several kernel functions $\kappa$ are tested, such as the Gaussian, polynomial, and
Matern~\cite{williams2006gaussian}, among which Gaussian kernels are found to produce generally lower NRMSE values.
Hyperparameters are identified by grid search: $\lambda_1, \lambda_2, \lambda_l$, $\lambda_u \in [10^{-5}, 1]$
in~\eqref{eq:graph.task.loss}; $N_{\mathit{l}}=10l$ for $l\in \llbracket 5, 15\rrbracket$;
$\vect{r}_1=(1,8r_1,8r_1,\ldots, 8r_1,1)$ and $\vect{r}_2=(1,8r_2,8r_2,\ldots,8r_2,1)$ for $r_1, r_2 \in \llbracket 1,12
\rrbracket$; $m \in \llbracket 1,2 \rrbracket$.

\begin{figure}[!t]
  \centering
  \resizebox{\linewidth}{!} { \begin{tikzpicture}
\begin{axis}[
    ybar,                         
    bar width=12pt,                 
    width=18cm, height=8cm,        
    enlarge x limits=0.05,         
    ylabel={Time (minutes)},
    xlabel={Methods},
    xlabel style={yshift=9pt},
    symbolic x coords={M1,M2,M3,M13,M4,M5,M6,M7,M8,M9,M10,M11,M12},
    xtick={data},
    xticklabels={{HodgeNet},{RTTC},{STTC},{NCP},{$(1,1)$},{$(1,2)$},{$(1,3)$},{$(2,1)$},{$(2,2)$},{$(2,3)$},{$(3,1)$},{$(3,2)$},{$(3,3)$}}, 
    xticklabel style={rotate=30, yshift=5pt}, 
    ymajorgrids = true,
    ymin=0,ymax=30,
    font=\Large,
    legend style={at={(0.02,0.98)}, anchor=north west}
]

\addplot coordinates {
  (M1,12.57) (M2,2.7) (M3,11.6) (M13,14.51) (M4,2.753) (M5,7.620)
  (M6,9.100) (M7,6.767) (M8,7.423) (M9,18.940) (M10,13.78) (M11,17.08) (M12,24.713)
};

\addplot coordinates {
  (M1,14.983) (M2,8) (M3,15.43) (M13,15.53) (M4,3.61) (M5,7.68)
  (M6,11.4) (M7,6.09) (M8,9.69) (M9,21.58) (M10,28.31) (M11,26.76) (M12,28.76)
};

\legend{EMA, BF}

\end{axis}
\end{tikzpicture} }
  \vspace{-0.5cm}
  \caption{Average computation times (minutes) across all sampling ratios vs.\ employed methods (with hyperparameters
    achieving the lowest NRMSE). Time is measured until a stopping criterion, similar to the one in
    line~\ref{alg.step:stop} of \cref{alg:multil.general} with $\epsilon = 10^{-4}$, is satisfied. The notation $(P, Q)$
    refers to \krettah with $P, Q$ corresponding to the Hadamard products in~\eqref{eq:model}. Times for
    PS~\cite{roddenberry2023signal} and S-VAR~\cite{krishnan2024simplicial,money2024evolution} are not displayed ($< 1$
    minute).}
  \label{fig:time}
\end{figure}
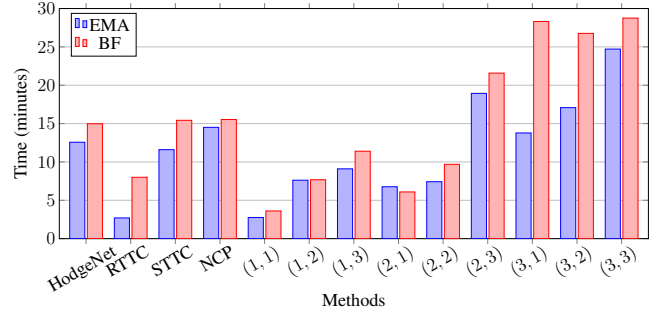

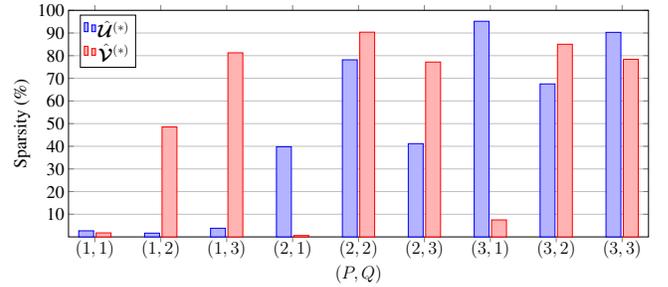
\begin{figure}[!t]
  \centering
  \resizebox{\linewidth}{!} { \begin{tikzpicture}
\begin{axis}[
    ybar,                         
    bar width=12pt,                 
    width=18cm, height=8cm,        
    enlarge x limits=0.05,         
    ylabel={Sparsity (\%)},
    ytick={0.1, 0.2, 0.3, 0.4, 0.5, 0.6, 0.7, 0.8, 0.9, 1},
    yticklabels={10, 20, 30, 40, 50, 60, 70, 80, 90, 100},
    ymajorgrids = true,
    xlabel={{$(P,Q)$}},
    symbolic x coords={M4,M5,M6,M7,M8,M9,M10,M11,M12},
    xtick={data},
    xticklabels={{$(1,1)$},{$(1,2)$},{$(1,3)$},{$(2,1)$},{$(2,2)$},{$(2,3)$},{$(3,1)$},{$(3,2)$},{$(3,3)$}}, 
    xticklabel style={rotate=0, yshift=5pt}, 
    ymin=0,ymax=1,
    font=\Large,
    legend style={at={(0.02,0.98)}, anchor=north west}
]

\addplot coordinates {
  (M4,0.0268973214285714) (M5,0.01659598214285714)
  (M6,0.03819196428571429) (M7,0.397963169642857) (M8,0.7815401785714285) (M9,0.4116685267857143) (M10,0.951841517857142) (M11,0.675256696428571) (M12,0.90296875)
};

\addplot coordinates {
  (M4,0.017732997481108313) (M5,0.48544080604534007)
  (M6,0.812544080604534) (M7,0.006171284634760705) (M8,0.90367758186397969) (M9,0.7718136020151134) (M10,0.07496221662468514) (M11,0.8503778337531486) (M12,0.7838790931989924)
};

\legend{{$\hat{\vectcal{U}}^{(*)}$}, {$\hat{\vectcal{V}}^{(*)}$}}

\end{axis}
\end{tikzpicture} }
  \vspace{-0.5cm}
  \caption{Average sparsity of $\hat{\vectcal{U}}^{(*)} \coloneqq \odot_{p=1}^P
    \hat{\vectcal{U}}_p^{(*)}$ and $\hat{\vectcal{V}}^{(*)} \coloneqq \odot_{q=1}^Q
    \hat{\vectcal{V}}_q^{(*)}$ across all sampling ratios. The sparsity is measured as the percentage of
    tensor entries with absolute values $\leq 10^{-3}$ after normalizing all entries by the largest-magnitude entry. }
  \label{fig:sparsity}
\end{figure}

\cref{fig:nrmse} reports NRMSE across sampling ratios. \krettah achieves the lowest errors except in the BF network at $s=0.1$, where it ranks second to PS~\cite{roddenberry2023signal}. HodgeNet~\cite{rodd2019hodgenet} performs the second-best in the EMA dataset but degrades sharply in the BF one, while S-VAR~\cite{krishnan2024simplicial,money2024evolution} exhibits the highest errors across all ratios due to reliance on incomplete past data.
\cref{fig:nrmse.sen} reports the effect of $(P,Q)$ on NRMSE, with $(1,2)$ consistently best in both datasets, supporting the use of HP.
\cref{fig:time} further underscores the efficiency of \krettah: at the lowest-NRMSE setting $(P,Q) = (1,2)$, the computation time is lower than that of HodgeNet~\cite{rodd2019hodgenet}, STTC~\cite{yu2025robust}, and NCP~\cite{bazerque2012nonparametric}, and comparable to RTTC~\cite{steinlechner2016riemannian}. As seen
in~\cref{fig:sparsity}, larger $P$ and $Q$ increase sparsity. Notably, the $(1,2)$
setting achieves the lowest NRMSE while reducing parameter storage by more than \num{40}\%.



\clearpage 


\onecolumn
\raggedcolumns 
\begin{multicols}{2}
\printbibliography[title={\normalsize\uppercase{References}}]
\end{multicols}

\end{document}